\documentclass[letterpaper, 10 pt, conference]{ieeeconf}  % Comment this line out if you need a4paper

\IEEEoverridecommandlockouts                              % This command is only needed if 
\usepackage{times}
\usepackage{graphicx}
\DeclareGraphicsExtensions{.pdf,.png,.jpg}
\usepackage{amsmath}
\usepackage{amssymb}  % assumes amsmath package installed
\usepackage{makecell}
\usepackage{booktabs}
\usepackage{multirow}
\usepackage{here}
\usepackage{kotex}
\usepackage{makecell}

\newcommand{\specialcell}[2][c]{%
	\begin{tabular}[#1]{@{}c@{}}#2\end{tabular}}
\title{\LARGE \bf
	Anchor Distance for 3D Multi-Object Distance Estimation from 2D Single Shot 
}

\author{Hyeonwoo Yu and Jean Oh% <-this % stops a space
	\thanks{Hyeonwoo Yu and Jean Oh are with the Robotics Institute of Carnegie Mellon University, Pittsburgh, PA 15213, USA
		{\tt\small \{hyeonwoy,hyaejino\}@andrew.cmu.edu}}%
}

\begin{document}
	
	\maketitle
	\thispagestyle{empty}
	\pagestyle{empty}
	
	%%%%%%%%% ABSTRACT
	\begin{abstract}
		%We present a real-time distance estimation method for multi-object using anchor distance.
		%	자율 주행이나 모바일 로봇을 위한 SLAM에서, 물체의 semantic feature뿐만 아니라 3차원 위치 또한 중요하다.
		%Semantic features and 3D location of object are key factors of autonomous driving and simultaneous localization and mapping (SLAM).
		%Understanding the semantic features and 
		Visual perception of the objects in a 3D environment is a key to successful performance in autonomous driving and simultaneous localization and mapping (SLAM).
		In this paper, we present a real time approach for estimating the distances to multiple objects in a scene using only a single-shot image.
		%	2D image로부터 얻은 2d detection을 이용하여 distance를 직접 계산할 수 있으나, 2d detection의 uncertainty와 computation burden 때문에 실시간 시스템에 적용하기가 힘들다.
		%Although a 3D distance to an object can be calculated directly using the detections made on a 2D image combined with the object parameters, it is challenging to apply such methods to a real-time system due to the uncertainty and the computational burden.
		Given a 2D Bounding Box (BBox) and object parameters, a 3D distance to the object can be calculated directly using 3D reprojection; however, such methods are prone to significant errors because an error from the 2D detection can be amplified in 3D.
		In addition, it is also challenging to apply such methods to a real-time system due to the computational burden.
		%	기존의 multi-object detection 기법의 경우 2d image 위에서의 2d BBox 추정이나 segmentation에 mostly pay attention하고 있다.
		In the case of the traditional multi-object detection methods, %they mostly pay attention to 
		existing works have been developed for 
		specific tasks such as object segmentation or 2D BBox regression.
		%	네트워크는 정교한 2d BBox 추정을 위해 anchor BBox를 도입하고, predictor들은 특정 2d BBox에 특화되어 학습되므로 distance에 대한 variation을 학습하기가 힘들다.
		These methods introduce the concept of anchor BBox for elaborate 2D BBox estimation, and predictors are specialized and trained for specific 2D BBoxes.
		%	우리는 2D image로부터 3D object의 distance 추정이 목적이므로, distance를 기준으로 하는 anchor distance를 정의하여 muti-object detection structure에 적용하는 기법을 제안한다.
		%Since our goal is to estimate the distances to the 3D objects from a single 2D image, we define anchor distance by object's location and propose a method that applies the anchor distance to the multi-object detector structure.
		In order to estimate the distances to the 3D objects from a single 2D image, we introduce the notion of \textit{anchor distance} based on an object's location and propose a method that applies the anchor distance to the multi-object detector structure.
		%	이로써 정교한 distance 추정과 동시에 실시간 수행을 achieve할 수 있다.
		%In this way, it is feasible to achieve the precise distance estimation and real-time execution simultaneously.
		%	anchor distance를 이용해 predictor에 prior를 주고, 네트워크 학습 시에도 distance를 기준으로 predictor가 특정 distance의 추정에 특화될 수 있도록 한다.
		We let the predictors catch the distance prior using anchor distance and train the network based on the distance.
		The predictors can be characterized to the objects located in a specific distance range.
		%	다양한 format의 distance를 이용해 anchor distance를 정의하고 네트워크를 학습할 수 있으며, 실제 실험을 통해 각 format의 distance의 효용성을 검증하였다.
		%Various distance formats such as log-scale or squared can be exploited to define anchor distance.
		By propagating the distance prior using a distance anchor to the predictors, it is feasible to perform the precise distance estimation and real-time execution simultaneously.
		The proposed method achieves about 30 FPS speed, and shows the lowest RMSE compared to the existing methods.
		% 	We evaluate the effectiveness of each anchor distance.
	\end{abstract}

	\section{Introduction}
	%로보틱스에서, 실시간 visual environment perception은 다양한 작업을 수행하기 위해서 필수적이다 such as SLAM and autonomous driving.
	%특히, With the emergence of neural networks, 보다 정확한 물체 인지를 위한 다양한 연구가 진행되었으며 물체의 다양한 disentangled features such as shape, orientation or dimensions of 3D BBox에 대한 추정이 가능하게 되었다.
	%물체를 정확히 인지하는것과 마찬가지로, 물체의 3d coordinate를 추정하는 것 또한 SLAM을 통해 로봇의 정확한 위치 추정과 object를 포함한 layered mapping을 수행하거나 obstacle avoidance를 수행하는 데 중요하다.
	Real-time visual perception and understanding of an environment is critical to the successes of robotic applications including autonomous driving.
	Visual Simultaneous Localization and Mapping (SLAM), in particular, is essential for performing navigation and exploration tasks as well as supporting high-level missions that require reliable mobility.
	Recently, SLAM focuses on two major facets: semantic recognition and spatial understanding \cite{slam++,categorySpecificSLAM,yang2019cubeslam,bowman2017probabilistic,multimodalSLAM}.
	With the advancement of deep neural networks, several approaches have been developed to achieve highly accurate results on semantic recognition, taking advantage of rich and sophisticated semantic features and various disentangled features such as shape, orientation, or dimensions of 2D and 3D Bounding Box (BBox) \cite{marrnet,myIROS2018,myICRA2018,dataDriven3Dvoxel,3dBBoxdeeplearninggeo,disentanglingrepresentations}.
	In this paper, we focus on real-time spatial understanding including accurate estimation of object locations specifically addressing the challenge in a monocular camera setting.
	
	%대부분의 자율주행이나 SLAM을 위한 robotic system에서는 물체의 거리를 구하기 위해 기존의 inverse reprojection method를 많이 사용한다.
	%물체의 3D shape이나 3d BBox를 image의 segmentation 또는 2d BBox에 reprojection하여 최적의 distance를 구한다.
	%그러나 이러한 기법은 image segmentation 또는 2D BBox estimation 결과에 굉장히 민감하다.
	%따라서 물체가 image의 side에 위치하거나 거리가 멀어서 2d detection이 정확하게 되지 않을 경우,
	%거리 추정 결과는 큰 오차를 지니게 된다.
	%또한 물체의 개수에 따라 linear하게 연산량이 늘어나기 때문에 multi-object detection에서는 실시간으로 동작하기가 어렵다.
	The majority of existing work on SLAM systems for autonomous driving relies heavily on an inverse reprojection method \cite{mousavian20173d,frost2016object,frost2018recovering,sucar2018bayesian} where an optimal distance is computed by reprojecting a 3D shape or its 3D BBox to a corresponding 2D BBox.
	The performance of such a method, however, tends to be extremely sensitive to that of image segmentation and/or 2D BBox estimation.
	%For instance, when an object appears in peripheral locations of an image or when the distance is too far from the camera to make a correct 2D BBox regression, these conditions can result in significant errors in 2D BBox inference and distance estimation consecutively.
	%In addition, this approach is not recommended for real-time multi-object detection such as autonomous driving in urban environments since the computational burden cumulates as the number of objects increases.
	\begin{figure}[t]
		\centering%
		\includegraphics[scale=0.10]{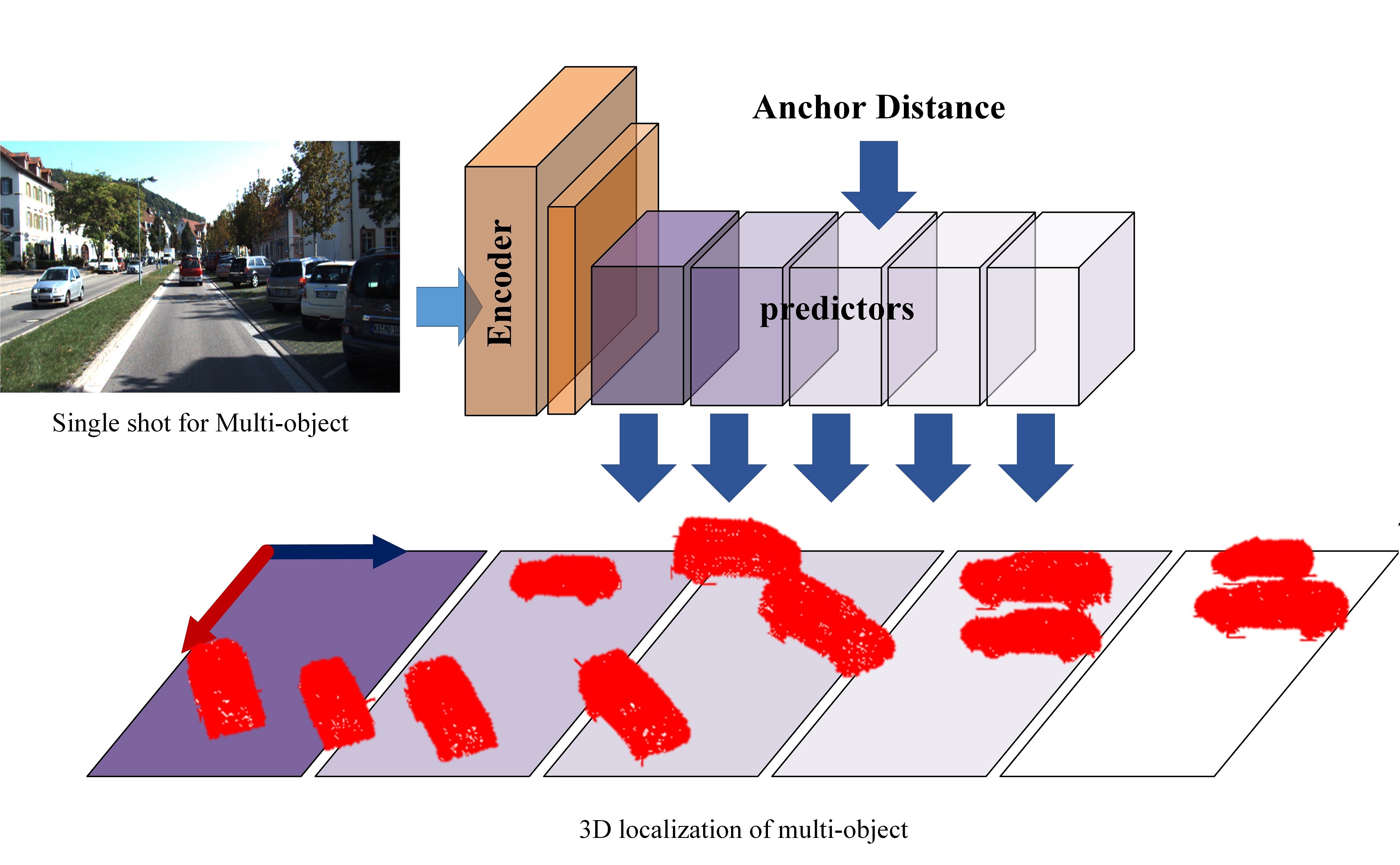}%
		\caption{%
			The overview of the proposed method.
			The network estimates the distance of multi-object with multiple predictors from single image.
			Each predictor is specialized to the object located in specific distance range by using anchor distance.
			With the anchor distance, predictors are provided priors of the distance so that accurate estimation under simple and fast network structure can be achieved.%
		}%
		\label{method_overview}%
		\vspace{-6pt}
	\end{figure}
	% RGB-dept estimation 기법은 single image의 3차원 spatial information을 더욱 풍부하게 해준다.
	% 추정한 depth map과 물체의 2D bbox를 이용하여 물체의 3차원 좌표를 추정하는 경우, bbox 내의 median depth를 이용한다.
	%따라서 highly occluded되어 있는 물체의 경우 outlier pixel를 제거하기 힘들어진다.
	% 물체의 segmentation을 이용하더라도, 계산 부담이 클 뿐만 아니라 물체의 partial depth로부터 3d center의 depth를 정확하게 계산하기 힘들다.
	
	Alternatively, depth estimation methods~\cite{fu2018deep,facil2019cam,wofk2019fastdepth} can be utilized for object-level 3D estimation, by taking the median depth of all geometric features or pixels within the detected 2D BBox~\cite{bowman2017probabilistic,frost2016object,zhong2018detect}.
	In these methods, however, it is hard to discard the outlier pixels especially if an object is severely occluded.
	%Even with the segmentation methods such as~\cite{maskrcnn}, which can add significant computational burden for minor improvements, estimating the depth of object's center from depth values of partial observation is still challenging.
	Adding a segmentation method such as~\cite{maskrcnn} can have minor improvements in the performance at the cost of significant computational burden; estimating the depth of object's center from depth values of partial observation is still challenging.
	%최근 single shot에 대한 multi-object detection 기법이 다수 연구되어 실시간 robot performance에 적용할 수 있게 되었으나, 이 기법들은 물체의 category classification, segmentation and 2d BBox detection에 특화되어 있다.
	
	To achieve robust object depth estimation, we can exploit existing single-shot, multi-object detection approaches, which are specialized in category classification, segmentation, and 2D BBox regression \cite{maskrcnn,redmon2017yolo9000,SSD,fasterRCNN}.
	%최근에 물체의 다양한 disentangled representation을 학습하기 위한 다양한 기법들이 제안되었다.
	Based on such methods, various approaches have been proposed for object disentangled representation \cite{seamlessSingleShot6D,voxelNet_yolo3D,3D-RCNN,huo2020learning}.
	%대부분의 3d multi-object detection methods using single image는 multi-object detector를 이용한 후, post-processing을 수행한다.
	Some works on multi-object understanding use multi-object detector to obtain object region followed by a post-processing step~\cite{zhu2019learning,dataDriven3Dvoxel,simonelli2019disentangling}.
	%혹은 Region proposal network (RPN)을 backbone과 함께 학습하여 Region of Interest(RoI) pooling을 수행하여 visual feature을 얻는다.
	Others train their baseline network with Region Proposal Network (RPN) and perform Region of Interest (RoI) pooling to obtain visual features of objects \cite{zhang2020regional}.
	%이에 더하여 물체의 representation을 정확하게 학습하기 위해, 별도의 estimation network module을 따로 학습한다.
	These methods deploy additional structure to estimate various object representations in addition to the baseline for multi-object detection.
	%따라서 네트워크의 복잡도가 증가하여 실시간 수행이 힘들다.
	%Hence, the complexity of the network increases and achieving real-time performance goes challenging.
	Such an increased complexity of the network makes real-time performance challenging~\cite{maskrcnn,zhu2019learning}.
	%네트워크의 구조를 단순하게 하기 위해, 물체에 대한 prior knowledge를 줄 수 있다.
	To simplify the structure of the network, prior knowledge of an object can be exploited.
	%기존의 multi-object detector는 2d BBox를 정확하게 추정하는 것이 목적이기 때문에, 2d BBox에 대한 prior로 anchor BBox을 사용한다.
	Since the purpose of the existing multi-object detector is to perform the 2D BBox regression, anchor BBoxes are used as prior knowledge of 2D BBox \cite{fasterRCNN,SSD,bochkovskiy2020yolov4}.
	%anchor BBox를 이용하여 network에게 object에 대한 prior knowledge를 주기 때문에, network 또한 anchor BBox에 따라 multiple predictor를 두어 학습할 수 있다.
	To utilize such prior knowledge represented by anchor BBoxes, several methods construct their networks by deploying multiple predictors according to anchor BBoxes.
	%이러한 기법들은 물체의 2d BBox on image를 기준으로 하여 network를 학습시키기 때문에, 물체의 정교한 거리 추정을 목적으로 기법을 적용하는 데 한계가 있다.
	These methods, however, have limitations in learning object representations such as distance estimation as they mainly focus on the 2D BBox on the image plane.
	
	%본 논문에서는 multi-object detector의 real-time performance와 정확한 distance 추정을 모두 얻기 위한 방법을 제안한다.
	%2D image로부터 3D space에서의 object distance를 정확하게 추정하는 것이 목적이므로, 물체의 distance를 이용하여 anchor distance를 정의한다.
	%predictors 또한 anchor distance를 이용해 물체의 distance에 대한 prior를 제공받는다.
	%predictors는 각기 다른 범주의 distance에 특화되어 학습하게 되므로, 추정하는 값에 대한 variation이 적어져 더 정교한 추정이 가능해진다.
	%기존의 multi-object detector가 anchor BBox를 얻기 위해 BBox의 IoU, area 차이 또는 dimensions의 차이 등을 이용할 수 있다.
	%마찬가지로 anchor distance 또한 normal, squared 또는 log-scale form을 이용할 수 있다.
	%%학습시에도 anchor distance를 얻을 때의 distance form을 이용하여 물체를 predictor에 assign한다.
	To bridge the gap between existing multi-object detection and distance estimation, we introduce the notion of~\textit{anchor distance}.
	%The overview of the proposed method is displayed in Fig.~\ref{method_overview}.
	We then propose a multi-object distance estimation method specifically designed for both real-time performance and accurate estimation.
	Given a 2D image as an input, we aim to estimate the distance to objects in a 3D space.
	Our work makes the following contributions to the state-of-the-art methods:
	\begin{itemize}
		\item as shown in Fig.~\ref{method_overview}, we transform the 2D single shot to 3D spatial grids represented as predictors so that the proposed method achieves real-time performance without a 2D RoI pooling network;
		\item compared to existing 2D detectors, the proposed method achieves robust detection results with overlapped objects since objects are detected in 3D grid space; and 
		\item we define and utilize the notion of anchor distance, thus predictors in our proposed network are specialized and robust to the objects in specific distance range.
	\end{itemize}
	% In order to achieve this goal, we define an anchor distance based on the distance (or size) of an object.
	% Using this anchor distance, the predictors receive prior knowledge of the object distance.
	%Using the anchor distance defined based on the distance to an object, the predictors receive prior of the object distance.
	%In this way, the predictors are trained to learn the distance for each specific distance range, resulting in a model that can accurately estimate distance with significantly reduced variances.
	%In addition to anchor distance, the proposed idea can also be generalized to using normal, squared, or log-scale forms.
	%Existing multi-object detectors use the IoU, the difference in area or the difference in dimensions of the 2D BBox to obtain the anchor BBox.
	%The anchor distance can be defined using normal, squared or log-scaled distance of objects.
	%The proposed method shows over 30 FPS which is about $4$ times faster than state-of-the-art method, and shows competitive results in Abs Relative (Abs Rel) difference, and outstanding results in RMSE.
	When compared to the state-of-the-art method, the proposed method runs about $4$ times faster at 30 FPS and shows competitive results in Abs Relative (Abs Rel) difference, and outstanding results in RMSE.

	\section{Related work}
	%% slam을 위한 물체 거리 얻는 기법
	%% multi-object detection - anchor box 쓴다
	%% 3d object detection 또는 distance 기법 - seamless orientation 기법, anchor box 쓰는거 잇는데 BBox기준임
	%% 나는 anchor distance 쓴다
	%모바일 로봇의 obstacle avoidance나 object-oriented slam 또는 자율 주행을 위해서, 관측한 물체의 3d 위치 추정은 중요하다.
	%In robotics, 대부분의 monocular SLAM using object detection는 distance estimation을 위해 traditional inverse perspective projection을 이용할 수 있다.
	%The 2D projection of an object is invertible since its extent parameters resolve depth ambiguity.
	%For example, object detection algorithm을 통해 얻은 2d BBox와 물체의 3d BBox parameters를 이용하여 물체의 거리를 구할 수 있다.
	%또는 autonomous driving과 같이 특수한 상황에서, 카메라의 높이를 이미 resulting hyper-parameter로 둘 수 있다. 그러면 모든 물체들이 바닥에 놓여있다는 강력한 가정을 통해 scale ambiguity를 해결하고 물체의 위치를 구할 수 있다.
	%그러나 이러한 기법들은 2d BBox estimation을 위한 추가적인 알고리즘이 반드시 필요하다.
	%또한 2D projection에 기반한 기법의 특성상, 물체의 위치 추정 결과가 2d BBox의 정확도(정교함)에 크게 좌우된다.
	% compulsory
	% mentadory
	% obligatory
	In the context of mobile robot navigation, 3D object detection and localization are compulsory to perform collision avoidance, object-oriented SLAM or safe navigation in autonomous driving.
	In this section, we discuss related works on monocular SLAM specifically focusing on the 3D localization.
	Various existing works on this end adopt the idea of inverse perspective projection \cite{frost2016object,frost2018recovering,yang2019cubeslam,sucar2018bayesian}. 
	The 2D projection of an object is invertible since its extent parameters resolve depth ambiguity. 
	That is, by mapping between 2D BBoxes from an object detection module and 3D BBoxes, these approaches can estimate the distance in 3D given a 2D input image.
	Also, in a special setting such as autonomous driving, we can assume that the height of a camera is known and all of the objects are placed on the ground.
	This allows the algorithm to resolve the scale-ambiguity and estimate object location.
	These approaches are still based on 2D BBox regression, the quality of distance estimation is also bounded by the precision of the 2D BBoxes.
	
	%In order to obtain 2D perception such as 2D BBox regression, it is inevitable to use additional object detection method.
	%최근 convolutional neural network를 이용한 multi-object detection 기법의 발달로, 실시간에 근접하게 동작할 수 있는 알고리즘이 다수 제안되었다.
	%They mainly focus on popular tasks such as detection using 2d BBox regression, object categorization and segmentation.
	%기존의 multi-object perception은 2D single shot을 타겟으로 하므로 2d BBox regression의 정확도가 곧 detection의 정확도의 척도가 되었다.
	%따라서 보다 정교한 BBox regression을 위한 테크닉과 네트워크 구조들이 제안되었다.
	%다양한 크기의 BBox를 보다 정확히 추정하기 위해, anchor box를 정의하여 BBox의 대략적인 크기에 대한 prior를 네트워크에 전달하는 기법이 사용된다.
	%각 anchor box는 네트워크의 서로다른 predictor에 assign된다.
	%각 predictor는 특정한 크기의 BBox를 추론하는 데 특화되어, 네트워크의 부담을 줄이는 동시에 더욱 정교한 BBox regression이 가능하게 된다.
	To obtain 2D perception such as 2D BBox regression, using an additional object detection method is inevitable.
	Several convolutional neural network (CNN)-based algorithms have been proposed to perform multi-object detection in real time.
	These approaches are mainly customized for specific tasks such as detection using 2D BBox regression, object categorization, and segmentation.
	Because the performance of these approaches depends on the quality of 2D BBoxes inferred from an object detector, results of 2D BBox detection have been commonly used as an evaluation criterion for the overall performance \cite{fasterRCNN}.
	This trend has led to the development of new techniques and network designs to improve the performance of 2D BBox regression \cite{SSD,bochkovskiy2020yolov4}.
	
	%기존의 multi-object perception 기법은 BBox regression 및 category estimation에 집중하였으나, 물체의 orientation, shape 또는 distance 등 다양한 disentangled representation을 추정하는 기법들이 제안되었다.
	Existing multi-object perception techniques mainly focus on 2D BBox regression and category estimation, and their variations estimate disentangled representations such as orientation, shape, or distance of objects \cite{myIROS2018,seamlessSingleShot6D,M3DRPN}.
	%이로써 single shot으로부터 3D multi-object detection을 할 수 있는 기법이 다수 연구되었다.
	%특히 distance estimation으로부터 물체의 3d coordinate를 추정할 수 있는 기법이 제안되었다.
	%제안된 많은 기법들은 object의 복잡한 disentangled representation을 학습하기 위해 visual feature을 이용한다.
	Most of the proposed methods exploit visual features to learn the complex representations of objects \cite{marrnet,disentanglingrepresentations,synthesizing3dshapes,image2mesh,liu2020smoke}.
	%따라서 region of interest (RoI) pooling이 필수이므로 추가적인 multi-object detector를 사용하여 2d BBox를 제공받아야 한다.
	Therefore, by using an additional multi-object detector, 2D BBox is provided for RoI pooling in order to obtain visual features for the target region.
	%기존의 multi-object detector structure을 변형하여 object understanding을 directly 학습하는 기법이 제안되었다.
	Other methods are proposed by modifying the existing multi-object detector structure for directly extracting features and performing object understanding \cite{simony2018complex,seamlessSingleShot6D,maskrcnn,zhang2020regional}.
	%그러나 여전히 RoI pooling을 위한 Region proposal network를 함께 학습한다.
	However, they still deploy the RPN for RoI pooling.
	%또한 물체의 depth 또는 distance와 같이 variation이 큰 object representation을 학습하기 위해, detection network 이외에 object understanding을 위한 network를 더 구성하게 된다.
	Moreover, it is necessary to construct the networks in addition to the baseline structure for object understanding tasks with a large variation such as depth estimation.
	%이 외에도 object의 non-linearity를 커버하여 정교한 추정을 수행하기 위해, they use networks with drawbacks such as increase of memory footprint.
	For these networks they use structures with several drawbacks such as the increase of memory footprint to cover the non-linearity of object understanding and perform elaborate estimation \cite{zhang2020regional,M3DRPN}.
	%이 방법들은 물체의 다양한 요소들을 representation하는 데에는 효율적이나, 그 복잡한 구조로 인해 실시간으로 수행하는 로봇에 적용하기 힘들다.
	These methods are effective in representing specific aspects of objects, but it is challenging to apply to mobile robot systems in real-time due to their complexity.
	
	% anchor BBox사용한게 있지만 다 BBox중심
	%네트워크의 부담을 줄이고 더 정확한 estimation을 위해, anchor를 이용하여 object에 대한 prior information을 네트워크에 제공하는 기법이 다수 제안되었다.
	In order to relax the network and achieve more robust results, a number of methods have been introduced to provide prior information about objects to the networks using anchors.
	%초기에 2D detection을 위한 기법들은 anchor BBox을 활용하였으며, 3D object detection을 위한 기법들은 anchor에 2D BBox뿐만 아니라 orientation이나 distance등을 include한다.
	Techniques for 2D detection exploit anchor of 2D BBox \cite{fasterRCNN,SSD,bochkovskiy2020yolov4}, and techniques for 3D include orientation and distance as well as 2D BBox in anchor \cite{M3DRPN,huo2020learning}.
	%그러나 anchor들은 여전히 물체의 2D BBox를 기준으로 한 clustering으로부터 정의된다.
	However, in these methods, anchors are still defined by clustering based on the 2D BBox of object on 2D image plane.
	%물체의 다양한 representations such as distance는 물체의 2D BBox와 directly proportional하지 않기 때문에, localization을 학습하기 위해 사용할 anchor로는 적합하지 않다.
	Since various object representations such as locations, shape, or orientation are not directly proportional to the projected 2D BBox, it is not suitable to use anchors arranged with 2D BBox for learning the distance of objects.
	%따라서 우리는 object 3D localization을 위한 anchor distance를 정의하고, 이를 이용하여 real-time performance를 보이는 네트워크를 학습하는 방법을 소개한다.
	Therefore, we define anchor distance for object 3D localization and introduce a method of training networks based on the anchors that achieves real-time performance.
	
	% \begin{figure}[t]
	% 	\centering
	% 	\includegraphics[scale=0.07]{./fig/distance_BBox_projection.jpg}
	% 	\caption{
	% 		An example of 2D bounding box projection.
	% 		The size of 2D bounding box is determined by the dimensions of 3D bounding box and location of the object.
	% 		Since the distance is sensitive to regression results of the 2D bounding box, small error of 2D box can cause enormous errors in distance estimation.
	% 	}
	% 	\label{distance_BBox_projection}
	% \end{figure}
	\section{Approach}
	%In robotics, estimating 3D coordinates of observed objects is necessary for various tasks such as SLAM.
	%로봇이 monocular camera로 물체를 관측하는 경우, 물체의 좌표를 추정하기 위해 image상의 2d BBox와 3차원 공간에서의 3d BBox의 dimensions 그리고 물체의 orientation를 이용할 수 있다.
	In general, when a monocular camera sensor is used for recognizing an object, the location coordinates can be estimated using the mapping between the 2D object detection and both the dimension and the orientation of a corresponding 3D BBox.
	% as illustrated in Fig.~\ref{distance_BBox_projection}. 
	%Assume that 2d BBox, dimensions of 3d BBox and orientations are given or estimated in advance.
	Assume that a 2D BBox of a detected object and the dimension and the orientation of a 3D BBox are given or have been estimated.
	%With the constraint that the 3D bounding box fits tightly into 2D detection box on 2d image, 3d coordinates of the object can be determined 유일하게.
	With the constraint that the 3D BBox fits tightly into the 2D detection box on the 2D input image, 3D coordinates of the detected object can be calculated;
	%그러나 estimating 3d coordinates by overlapping the projection of 3D BBox to 2d BBox usually causes inaccurate results due to the sensitivity of 2d BBox as shown in Fig.~\ref{distance_BBox_projection}.
	however, estimating 3D coordinates by overlapping the projection of 3D to 2D BBoxes usually causes inaccurate results due to the sensitivity to the size of the 2D BBoxes. % as shown in Fig.~\ref{distance_BBox_projection}.
	%It also takes significant calculation burden, as each side of the 2D BBox can correspond to any of the eight corners of the 3D BBox.
	%Even with the strong constraint that most of the objects are located on ground plane, at least 64 configurations should be checked.
	%Therefore, we propose the method based on network estimation to achieve more accurate and real-time performance as well.
	A small error in the size of BBox can cause a substantial error in the distance estimation calculation. 
	Furthermore, this approach can add a significant calculation burden, as each side of the 2D BBox can correspond to any of the eight corners of the 3D BBox.
	Even with the strong constraint that most of the objects are located on a ground plane, at least 64 configurations must be checked \cite{3dBBoxdeeplearninggeo}.
	To address these challenges, we propose an approach that achieves both high accuracy and real-time performance by directly estimating the object distance.
	%Instead of learning the mapping between 2D and 3D bounding boxes, the proposed approach uses the mapping between the center of the object in 2D and 3D as described in the next subsections.
	
	\subsection{3D Coordinate and 2D Bounding Box}
	Given 3D directions to the center of object and depth (or distance from origin), 3D coordinates can solely be determined.
	For estimating the center of the 3D object, the center of a 2D BBox can be exploited by reprojecting it to the 3D real-world coordinate.
	In this way, we can have a normalized ray direction vector toward the center of the 3D object from the origin.
	Hence, the 3D coordinate of an object can be obtained by multiplying this ray direction vector and estimated distance.
	%네트워크가 물체의 3d coordinate를 모두 직접 추정하도록 학습할 수도 있다.
	The network can also be trained to directly estimate all 3D coordinates of an object.
	%그러나 multi-object detector의 경우 grid 형태로 분포하고 있는 predictor을 활용하여 2d image상에서의 물체 중심을 비교적 정확히 추정할 수 있다.
	In case of using multi-object detector, however, it can be relatively advantageous to estimate the center of an object by using the predictors distributed in a grid form.
	%따라서 물체의 중심방향을 직접 reprojection하는 것이 x,y를 직접 학습하는 것보다 더 정확하다.
	Hence, it is more accurate to compute the coordinate from reprojection than learning the $x$,$y$, and $z$ components of location directly.
	%본 논문에서는 물체의 위치 추정을 위해 distance from origin과 2d BBox를 네트워크에 학습시킨다.
	We design our network to train on the distance from the origin and the 2D BBoxes as in \cite{redmon2017yolo9000,bochkovskiy2020yolov4}.
	
	%The main idea here is to use the 3D direction and the depth (the distance from the origin) to the center of object to estimate the 3D coordinates output.
	%For estimating the center of an object, the center of 2D bounding box can be exploited by reprojecting it to the 3D real-world coordinate.
	%In this way, we can have a normalized ray direction vector toward the center of the object.
	%The 3D coordinate of object can thus be obtained by multiplying this ray direction vector by the estimated distance.
	%%네트워크가 물체의 3D coordinate를 모두 직접 추정하도록 학습할 수도 있다.
	%%그러나 multi-object detector의 경우 grid 형태로 분포하고 있는 predictor을 활용하여 2D image상에서의 물체 중심을 비교적 정확히 추정할 수 있다.
	%%따라서 물체의 중심방향을 직접 reprojection하는 것이 x,y를 직접 학습하는 것보다 더 정확하다.
	%%본 논문에서는 물체의 위치 추정을 위해 distance from origin과 2D bounding box를 네트워크에 학습시킨다.
	%A network can be designed to learn to directly estimate the 3D coordinate of an object; however, in the case of a multi-object detector, it can be relatively advantageous to estimate the center of an object by using the predictors distributed in a grid form. That is, it is more accurate to compute the coordinate from reprojection than training the mapping in the network. We design our network to train on the distance from the origin and the 2D bounding boxes.  
	
	%여기서부터 영어로 ㄱ
	\subsection{Anchor Distance}
	In order to achieve the real-time multi-object distance estimation, our proposed method is based on a simple multi-object detection structure, namely, YOLOv2 \cite{redmon2017yolo9000}.
	%기존의 네트워크는 2d BBox의 dimension 밑 center만을 추정하기 때문에, 우리의 네트워크는 이에 더하여 물체의 distance from origin을 함께 추정한다.
	As the existing network only estimates the center and dimensions of a 2D BBox, we let our network predict the distance from the origin of the object as well.
	%그러나 기존의 multi-object detector는 2d image에서의 2d BBox 추정이 목적이었기 때문에, 각각의 predictors는 서로 다른 크기의 2d BBox를 전담하여 학습한다.
	The following problem, however, still remains:
	%the purpose of the conventional detector is to estimate not distance but 2D BBoxes so that each predictor is dedicated to learn for the corresponding BBoxes of different sizes.
	the purpose of conventional detectors is to estimate the 2D BBoxes rather than the distances to the objects, and each predictor are dedicated to learn for the corresponding BBoxes of different sizes.
	% %따라서 각 predictor가 추정해야 하는 2d BBox의 variation을 최대한 줄여, predictor의 부담을 줄여줌으로써 더욱 정확한 2d BBox 추정이 가능하게 한다.
	% %By reducing the variation of 2D BBoxes that each predictor should predict, the burden of the network is reduced during more accurate 2D BBox estimation.
	% To reduce such a burden of the network, these approaches reduce the variation of the 2D BBoxes that each predictor should predict, resulting in more accurate 2D BBox estimation. 
	% %이를 더욱 강화하기 위한 수단으로 anchor 2d BBox를 사용하여 2d BBox를 추정할 때 각 predictor에게 추정해야 하는 BBox size에 대한 prior knowledge를 준다.
	% %Therefore, 
	% Additionally, in order to achieve further improvement, a 2D anchor BBox is applied to each predictor as the prior knowledge about the BBox sizes.
	To reduce such a burden of the network, a 2D anchor BBox is applied to each predictor as the prior knowledge about the 2D BBox sizes.
	These approaches reduce the variation of the 2D BBoxes that each predictor should predict, resulting in more accurate 2D BBox estimation. 
	
	\begin{table}[t]
		\caption{Variance of the distance of 2D BBox groups and distance groups for KITTI dataset}
		\label{variance_of_distance}
		\begin{center}
			\begin{tabular}{c c c c}
				& & & (unit : $m^2$)\\
				\hline
				\hline
				\# of predictors  & \multirow{2}{*}{order}  & 2D BBox   & distance \\
				(groups) & &  grouping &   grouping \\
				\hline
				\hline
				\multirow{2}{*}{2} & 1 & 14.84 & 25.69 \\
				& 2 & 220.12 & 31.76 \\
				\hline
				\multirow{3}{*}{3} & 1 & 9.07 & 12.26 \\
				& 2 & 33.26 & 8.60 \\
				& 3 & 186.27 & 20.30 \\
				\hline
				\multirow{5}{*}{5} & 1 & 7.08 & 5.36 \\
				& 2 & 18.68 & 4.27 \\
				& 3 & 49.57 & 3.10 \\
				& 4 & 91.14 & 9.28 \\
				& 5 & 98.21 & 3.55 \\
				\hline
				\hline
			\end{tabular}
		\end{center}
	\end{table}
	%제안하는 기법은 물체의 3D coordinate를 정확하게 추정하는 것이 목적이므로, anchor box와 비슷하게 anchor distance라는 개념을 도입할 수 있다.
	%Since the proposed method aims to achieve 3D coordinate estimations of multi-object, the concept of anchor distance can be introduced which is similar to the case of anchor BBox.
	In order to achieve 3D coordinate estimations of multiple objects, we introduce the concept of anchor distance that is similar to the anchor BBox.
	%distance의 prior knowledge를 간단하게 얻기 위해, 각각의 anchor box에 속하는 물체들의 distance의 평균을 이용하여 각 anchor box에 대응하는 average distance를 정의할 수 있다.
	To obtain the prior knowledge of the distance in a simple manner, the average distance corresponding to each anchor BBox can be defined by using the average of distances of objects belonging to each anchor BBox group.
	%그러나 그림 [b]와 같이, anchor box의 크기는 물체의 distance와 비례하지 않는다.
	Unfortunately, the size of an anchor BBox is not exactly proportional to the distance of the object.
	%그러므로 distance를 anchor box에 귀속시키면 각 predictor는 특정 크기의 BBox 추정에는 특화되지만, 다양한 distance를 추정해야 하므로 predictor의 부담이 커지게 된다.
	%Therefore, when an anchor BBox simply includes the average distance, each predictor can be specialized in estimating 2D BBox in specific size, but the burden of distance estimation still remains because distances within a large range should be estimated from each predictor. In other words, the network still learns 2D BBox regression as the main task.
	When an anchor BBox simply includes the average distance, each predictor can undesirably learn the mapping between the size of a 2D BBox and a distance.
	In other words, the network still learns the 2D BBox regression as its main task, leaving the burden of distance estimation to each individual predictor where the distances within a large range should still be estimated.
	%
	% In other words, 네트워크는 여전히 2d BBox를 메인으로 학습하게 된다.
	
	%따라서 anchor BBox 대신 anchor distance를 정의하여 각 predictor들을 학습하도록 한다.
	%To bridge the gap between the detector and distance estimation,
	To address this issue, 
	we define the concept of an anchor distance to train each predictor as follows.
	%이와 같이 하면 각 predictor들은 특정 BBox에 특화되는 대신, 특정 거리의 물체들의 추정에 특화되어 정교한 거리 추정이 가능해진다.
	With an anchor distance, each predictor is specialized in estimating the distance of the object in a specific distance range, instead of being specialized in specific 2D BBoxes.
	%anchor BBox를 얻을때와 유사하게, 물체들의 distance들을 k-means clustering을 통해 그룹화하여 그 center들을 anchor distance로 정의한다.
	Similar to obtaining the anchor BBoxes, the distances of objects are grouped through $k$-means clustering.
	Each center of the groups (or clusters) is defined as anchor distance.
	%k=2,3,5일때, BBox로 cluster한 후 average distance를 구하였을 때의 variance와 distance로 clustering하여 anchor distance를 구하였을 때의 variance를 Table.~\ref{variance_of_distance}에 비교하였다.
	We compare the variance of the average distance obtained from 2D BBox clustering and that of the anchor distance from distance clustering for the case of $k=2,3,5$ in Table~\ref{variance_of_distance}.
	%KITTI dataset의 car category를 활용하였다.
	We use the \texttt{car} category in the KITTI dataset \cite{KITTI} as an example.
	%predictors의 order는 distance가 작은 것부터 정렬하였다.
	The predictors are sorted by the corresponding distance in ascending order.
	%2D BBox를 이용한 grouping의 경우, 먼거리를 위한 group의 variance가 가까운 거리를 위한 group의 variance보다 더 큰 것을 알 수 있다.
	In the case of 2D BBox grouping, the variance of the group for the long distance is greater than the one for the short distance.
	%물체가 멀리 떨어져 있을수록 2d BBox의 변화가 거의 없기 때문이다.
	This is because objects that are far away from the origin are similar in terms of size of the 2D BBox.
	%반면에 distance를 기준으로 clustering한 경우는, 2d BBox 를 이용하여 clustering하였을 때와 비교하여 variance가 훨씬 작은 것을 볼 수 있다.
	On the other hand, in the case of distance clustering, the variance is much smaller than that in the case of 2D BBox clustering for all predictors.
	%따라서 anchor distance를 사용하였을 때 predictors는 보다 일관적인 distance를 추론함으로써 더 정교한 추정 결과를 보일 수 있다.
	Therefore, predictors can show more precise estimation results as they infer more consistent distances with anchor distance.
	%anchor distance를 적용한 네트워크의 estimation 결과는 다음 식과 같이 나타낼 수 있다:
	The results of the network prediction using anchor distance are denoted as follows:
	\begin{align}
	d_i &= d^a_i \times exp(t_i)\text{ , }\text{ }i\in \{0,1,...,k-1\}
	\end{align}
	%$d = anchor_distance * exp(t)$
	%제안하는 네트워크의 overview를 그림에 나타내었다.
	where $t$ is an output of the network, and $d^a$ is the anchor distance.
	Similar to \cite{redmon2017yolo9000,bochkovskiy2020yolov4}, we use an exponential function as the final activation function of the output.

	%2d BBox는 distance estimation에는 영향을 주지 않지만, BBox center로의 ray direction을 구하여 3D coordinate를 계산하는 데 쓰이므로 여전히 중요하다.
	In our method, the dimension of 2D BBox has no effect on distance estimation directly, but the center of 2D BBox is crucial as it is exploited to calculate the 3D coordinate by finding the ray direction.
	%distance를 기준으로 얻은 각 clustering을 위한 BBox의 prior를 정의할 수 있다.
	To achieve the 2D BBox regression, priors of the BBox can be defined for our distance grouping.
	%BBox의 dimension을 산술평균 내어서 average BBox를 구할 수도 있다.
	Simply, we can define the average BBox of clusters for anchor distance by taking the arithmetic mean of dimensions of the BBoxes.
	%그러나 clustering은 유사한 distance에 초점이 맞춰져 있으므로 BBox는 좀 더 산발적으로 grouping 되어 있다.
	However, our clusters focus on distance so that BBoxes in a group have large variance in terms of their sizes.
	%따라서 보다 정확한 BBox average를 위해, dimension 차이가 최소가 아닌 IoU의 차이가 최소가 되는 average BBox를 구한다.
	For more accurate BBox regression with anchor distance groups, we take the average BBox approximately where we minimize not the differences of dimensions but the differences of intersection over union (IoU) as follows:
	\begin{align}
	\left(h^m\right)^2 = \frac{\sum w_j h_j^2}
	{\sum w_j}, \text{ }\text{ }
	\left(w^m\right)^2 = \frac{\sum h_j w_j^2}
	{\sum h_j}
	\end{align}
	where $h^m$ and $w^m$ are the height and width of the average BBox, respectively.
	$h$ and $w$ are height and width of BBox.
	Using this average BBox, the network output for BBox dimension is given as:
	\begin{align}
	\nonumber
	h_i &= h^m_i \times exp\left(u_i\right) \\
	w_i &= w^m_i \times exp\left(v_i\right)\text{ , }\text{ }i\in \{0,1,...,k-1\}
	\end{align}
	where $u$ and $v$ are outputs of the network.
	For the center of the 2D BBox, we follow the similar settings of \cite{redmon2017yolo9000}.
	%KITTI dataset에 대하여 실제 IoU로 구한 anchor BBox와 anchor distance에 따른 average BBox의 크기를 표에 비교하였다.
	%IoU구하는 식은 다음과 같다.
	%intersection은 다음과 같이 된다.
	%average BBox를 근사한 closed form 해를 구하기 위해서, IoU 구하는 식에서 intersection을 다음과 같이 근사한다.
	%그러면 average BBox는 다음과 같다.
	%이 식을 이용하여, anchor distance에 대응하는 average BBox를 구한다.
	\begin{figure*}[t]
		\centering
		\includegraphics[scale=0.41]{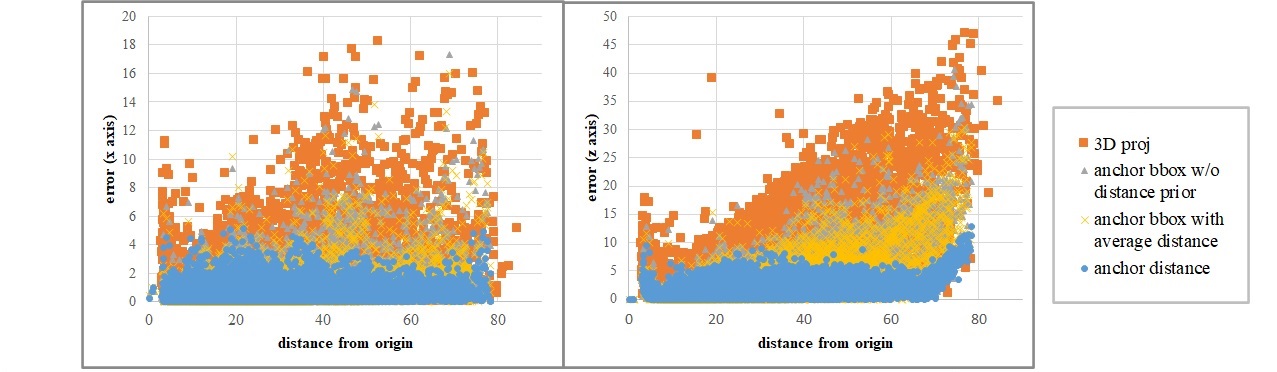}
		\caption{
			Distance estimation error of $x$ and $z$ axises, which denote the horizontal location and depth of the object.
			As the distance of the object increases, the estimation error increases in case of other methods.
			The proposed method shows consistent error independent to the distance from origin as the method exploits the anchor distance.	
		}
		\label{error_graph}
	\end{figure*}
	
	\subsection{Predictors}
	%anchor distance를 사용함으로써 네트워크에 object의 distance에 대한 prior를 제공해 줄 수 있다.
	With the anchor distance, we can provide the prior knowledge of the distance to our network.
	%각 anchor distance를 네트워크의 각 predictor에 assign하여, predictor가 특정 distance 근처에 존재하는 물체에 특화되어 학습될 수 있도록 한다.
	The predictor is specialized and trained for objects near specific distance as each anchor distance is assigned to each predictor of the network.
	%predictor가 추론해야 하는 distance 값의 variance가 낮아지면서, network의 부담을 줄이고 정확한 추론이 가능하게 한다.
	As the variance of the distance inferred by a predictor decreases, the complexity of the network is decreased.
	%네트워크의 predictor들은 3D region proposal network등의 별다른 추가적인 구조 없이도 자연스럽게 3차원 환경을 구성할 수 있게 된다.
	The predictors of the network can construct a 3D environment without additional structures such as 3D RPN or 3D convolutional layers.
	%본 논문에서는 YOLOv2 구조를 적용하여 네트워크를 구성하였다.
	
	%기존의 multi-object detector는 2D anchor BBox를 이용하였다.
	The existing multi-object detectors utilize 2D anchor BBoxes by clustering 2D BBoxes.
	%BBox를 clustering하기 위해 단순히 BBox의 dimensions를 이용하거나, area 또는 IoU의 차이를 이용할 수 있다.
	In order to cluster 2D BBoxes, we can use the dimensions of the BBox or IoU.
	%마찬가지로, anchor distance또한 다양한 format such as 단순 거리, squared distance or log-scale distance를 이용할 수 있다.
	Likewise, anchor distance can be achieved by using various formats such as normal, squared, or log-scaled.
	%본 논문에서는 세 가지의 경우 모두를 적용하여 anchor distance를 구한 후 네트워크를 학습하였다.
	In our work we apply all three cases to obtain the anchor distance and train the network.
	%anchor BBox를 이용한 기존의 multi-object detector는 학습할 때에도 predictor 중에서 target BBox와 가장 가까운 BBox를 추론한 predictor를 골라 target BBox를 assign한다.
	Whilst training, the existing multi-object detector using anchor BBox chooses the predictor that infers the closest BBox to the target BBox and assigns that target BBox to learn.
	%비슷하게, 제안하는 기법은 network의 predictor 중 target distance와 가장 가까운 값을 추론한 predictor를 골라 학습할 수 있다.
	Similarly, the proposed method learns the distance of the target object by selecting and training a predictor which infers the value closest to the target distance.
	%predictors가 추정한 distances와 target와의 차이를 계산할 때에도 anchor distance를 이용할 때의 distance format을 그대로 이용한다.
	Note that the same distance format used for obtaining anchor distance is also used when comparing the the difference between the target distance and estimated one by the predictor during training.
	
	%학습 초기에, variation이 큰 distance를 학습하기 위해 predictor의 추론값이 매우 크게 요동치는 것을 우리는 확인하였다.
	At the beginning of training, we found that the predictor's inference value highly fluctuates in order to learn distances with quite large variations.
	%predictors가 추정한 distance를 이용하여 학습할 object를 선택할 시, 네트워크가 일관되지 않게 학습되거나 object가 anchor distance에 무관하게 predictor에 assign된다.
	In this case, objects are inconsistently assigned to the predictors regardless of the anchor distance.
	%그러므로, 실제 학습 시에는 predictor 선택을 위해 predictor가 추정한 distance를 이용하지 않고 predictor의 anchor distance를 이용하였다.
	Therefore, during the training phase, we use the predictor's anchor distance rather than the estimated result for predictor selection.

	\subsection{Training Loss}
	%제안하는 기법에서 object의 distance from origin을 추정한다.
	In the proposed method, the distance from the origin of an object is estimated.
	%2D BBox를 이용하여 물체의 중심방향을 가리키는 ray vector을 얻을 수 있으므로, distance from origin 또는 depth를 이용하면 물체의 3D coordinate를 모두 알 수 있다.
	Since a ray vector indicating the center direction of an object is obtained using a 2D BBox, all 3 components of 3D coordinate can be obtained by using distance from origin or depth.
	%따라서 네트워크 학습 시에도 x,y and z coordinates를 따로 학습할 수 있다.
	Therefore, it is possible to separately train $x$,$y$, and $z$.
	%그러나 2d BBox와 물체의 3D location은 서로 독립이라고 가정하고, 2d BBox와 distance를 따로 학습한다.
	However, we assume that 2D BBox and 3D location are independent and let 2D BBox and the distance be learned separately.
	%2d BBox는 CIoU loss를 이용한다.
	For the 2D BBox training, we adopt CIoU loss \cite{zheng2019distanceiou,bochkovskiy2020yolov4}.
	%distance는 대체로 L1 loss를 주로 사용하나, 본 논문에서는 L2 loss를 사용하였다.
	For depth estimation, $L_1$ loss is generally used, but in our work $L_2$ loss is applied.
	
	\section{Implementation and Training Details}
	To implement the proposed observation model, we use the darknet19 structure \cite{redmon2017yolo9000} for the encoder backbone.
	We construct the encoder by adding 3 convolutional layers with 1,024 filters followed by one convolutional layer on top of the backbone.
	The final convolutional layer has
	$
	k \times 
	\left(
	4 + 1
	\right)
	$
	filters; $4$ for 2D BBox dimensions and center, and $1$ for the distance.
	%각각의 predictor는 물체의 BBox 및 distance from origin을 추정한다.
	Each predictor estimates 2D BBox and distance from the origin of the object.
	We pretrain the backbone network on the Imagenet classification dataset.
	We use the Adam optimizer with a learning rate of $10^{-4}$.
	For all training processes, a multi-resolution augmentation scheme is adopted.
	Similar to \cite{redmon2017yolo9000, myICRA2018}, Gaussian blur, HSV saturation, RGB invertion and random brightness are applied to 2D images while training.
	%We also vertically flip images and invert the sign of $x$ coordinate of the object in random.
	Random scaling and translation are not used in order to preserve the 3D coordinates of objects.
	
	\section{Experiments}
	% 제안하는 기법 predictor 개수 많을수록 정교해짐
	% 2d BBox와 3d BBox reprojection으로 계산한 거리 비교
	% 학습을 위해 같은 네트워크 3d BBox, orientation 학습
	% 2d BBox가 정교해질수록 잘되니까 predictor는 5개일때의 결과 비교
	% anchor BBox에 대응하는 average distance와 anchor distance 이용한 결과 비교 이때는 단순 distance format 이용
	% anchor distance 구할 때에 다양한 format 이용 가능 실험결과 비교
	% 3d visualization
	We evaluate the proposed method in various aspects.
	%기존 multi-object detectors와 유사하게, predictor의 개수가 많아질수록 task를 더욱 정교하게 수행할 수 있다.
	Similar to the traditional multi-object detectors, as the number of predictors increases, the estimation can be performed more precisely. 
	%따라서 모든 실험에서는 predictor 수를 다양하게, 예를 들어 2,3 그리고 5개, 하여 네트워크를 학습한다.
	In our experiments, we train the network with various numbers of predictors such as $2$, $3$, $5$ and so on.
	%기존 기법과의 비교를 위해 기존 방법을 포함한 몇 가지의 방법들을 적용하여 물체의 거리를 구하였다.
	We compare our methods to existing methods based on Faster R-CNN using RPN.
	A depth estimation method is also compared by using ground truth 2D BBox.
	We use the median depth within 2D BBox as the depth of a detected object.
	For the depth estimation method, we choose \cite{fu2018deep} since the method proposes ordinal regression in order to consider depth intervals.
	Additionally, we also implement several methods including ours in order to estimate the distance of objects -
	%우선 1) 2d BBox on 2d image에 3d BBox in 3d space를 projection하여 물체의 거리를 계산한 결과를 제안하는 방법의 결과와 비교하였다.
	1) \textit{3D proj}: we implement the method that calculates distance by projecting 3D BBox in 3D space to 2D BBox on 2D image.
	%이를 위해 제안하는 네트워크에서 distance를 추정하는 대신, 물체의 3d BBox의 dimension와 orientation을 추정한 후 기존 방법들과 유사하게 물체의 거리를 계산하였다.
	We construct the network which shares the same backbone with our proposed method, and train the network on dimensions of 2D, 3D BBox and orientations of objects.
	%2) 물체의 3차원 좌표를 모두 direct로 추론하는 것도 가능하다. 물체의 distance to origin을 추론하는 대신에 물체의 좌표 xyz를 모두 실수값으로 추론한 결과도 비교하였다.
	%다음으로 2) 기존의 multi-object detector와 유사하게 anchor BBox를 사용하면서, 각 그룹에 해당하는 물체의 average distance를 predictor의 prior distance로 이용하는 경우의 결과를 확인하였다.
	2) \textit{anchor BBox}: similar to the existing detector, we evaluate the method that uses anchor BBox and average distance of groups obtained from 2D BBox clustering.
	%average distance는 단순 산술평균으로 구하였다.
	In this case the average distance is calculated by taking arithmetic mean.
	We also evaluate the case without any distance prior.
	3) \textit{proposed method}: the proposed method using anchor distance and its training scheme is evaluated.
	%distance clustering을 수행할 때 distance format을 normal, squared 그리고 log-scale을 사용하였으며 각 format에 따라 모델을 학습하였다.
	For obtaining anchor distance by distance clustering, normal, squared, and log-scale formats are used for 3 individual models.
	For each model, the same format used for anchor distance is also used for training;
	while choosing the predictor for the target object, target distance and estimated distances from predictors are compared by using the same format of distance that is used for clustering.
	%학습 시에도 target object의 distance에 대해 가장 가까운 distance를 추정한 predictor을 선택할 때, clustering에서 사용한 format을 적용하였다.
	
	For all experiments, we use the \texttt{car} category in the KITTI 3D object detection dataset.
	Since each grid in our network has multiple predictors, for evaluation we choose the predictor which estimates the highest 3D IoU to the ground truth.
	
	\subsection{Relation Between Object Distance and Error}
	%우선 기존의 3D projection을 이용한 방법, anchor BBox without distance prior, anchor BBox with average distance and anchor distance를 Fig.~\ref{error_graph}에 비교하였다
	We compare the results of \textit{3D proj}, \textit{anchor BBox} with and without average distance, and the proposed method in Fig.~\ref{error_graph}.
	We plot the distance error of the object according to its distance from origin.
	%anchor BBox 및 anchor distance 기법은 $k=5$으로 설정하고, anchor distance는 squared format을 사용하였다.
	For the method \textit{anchor BBox} and the proposed method, we set $k=5$.
	For the distance format of the proposed method, we use squared format.
	Results of $x$ and $z$ axis are shown, which denote horizontal and depth of the object location respectively.
	We leave the result of $y$ axis out, since the height of cars have small variations compared to the depth of cars so that errors of $y$ axis are significantly smaller than errors of the other axes.
	%그래프에서 보듯이, 3d BBox를 2d BBox on image에 reprojection하여 얻은 결과는 물체가 먼 곳에 있을수록 정확도가 크게 떨어진다.
	As shown in the graph, the estimation result of \textit{3D proj} is significantly more inaccurate as the objects locate further.
	%물체가 멀리 있을수록 2d BBox의 크기가 작아지므로 정교하게 추론하기가 어려워져 3d BBox를 reprojection 시에 물체 위치에 대한 오차가 커진다.
	The further the objects are, the more similar the sizes of the 2D BBoxes are; therefore, it is challenging to infer the distance precisely since the estimations are highly sensitive to the errors of small 2D BBoxes of objects in a far distance.
	%anchor BBox를 사용한 방법 두 가지는 3D proj방법보다는 적은 오차로 물체의 distance를 추정한다.
	Compared to \textit{3D proj}, methods using anchor BBox have more robust inferences of distance.
	%그러나 원점으로부터 멀리 있는 물체일수록 $z$방향의 오차가 커진다.
	However, as the variations are high for the distance of objects located far from the camera, it is still challenging to estimate the far distance without a tremendous error.
	%이로부터 물체의 거리와 2d BBox가 정비례하지는 않으나 약간의 상관관계가 있음을 알 수 있다.
	From these results we can conclude that the distance and the 2D BBox of the object are not directly proportional to each other, but there exists a slight correlation.
	%anchor distance를 활용한 경우, 물체의 거리와 상관없이 대략적으로 균일한 추정 오차를 보인다.
	In the case of using anchor distance, estimations results show constant and uniform error regardless of the distance of the object.
	%거리가 멀어지더라도 anchor distance를 이용하기 때문에 거리에 상관없이 predictors가 추론하는 값의 variation이 일정하다.
	%Even though the distance of the object increases, the variation of the value inferred by predictors is almost consistent as the anchor distance gives a strong prior information.
	
	\begin{table*}[t]
		\caption{Variance of the distance of 2D BBox groups and distance groups for KITTI dataset}
		\label{distance_eval}
		\begin{center}
			\begin{tabular}{c |c| c |c c  c c c c c}
				\hline
				\hline
				\multirow{2}{*}{Method}  & \multirow{2}{*}{FPS} & \# of   & $\sigma < {1.25}$ & $\sigma < {1.25}^2$ &  & Abs Rel & Sqr Rel & RMSE & RMSE$_{log}$ \\
				& & predictors & \multicolumn{2}{c}{(higher is better)} && \multicolumn{4}{c}{(lower is better)} \\
				\hline
				SVR\cite{gokcce2015vision} &-& - & 0.345 & 0.595 & & 1.494 & 47.748 & 18.970 & 1.494 \\
				IPM\cite{tuohy2010distance} &-& - & 0.701 & 0.898 & & 0.497 & 35.924 & 15.415 & 0.451 \\
				\hline
				Zhu et al.(ResNet50)\cite{zhu2019learning} &\multirow{2}{*}{$<15$}& - & 0.796 & 0.924 & & 0.188 & 0.843 & 4.134 & 0.256 \\
				Zhu et al.(VggNet16)\cite{zhu2019learning} && - & 0.848 & 0.934 & & 0.161 & 0.619 & 3.580 & 0.228 \\
				\hline
				Zhang et al.(MaskRCNN[ResNet50])\cite{zhang2020regional} &\multirow{2}{*}{$<7$}& - & 0.988 & - & & 0.051 & - & 2.103 & - \\
				Zhang et al.(MaskRCNN[ResNet50] + addons)\cite{zhang2020regional} && - & \textbf{0.992} & - & & \textbf{0.049} & - & \textbf{1.931} & - \\
				\hline
				DORN (depth map estimation) \cite{fu2018deep} & - & - & 0.883 & 0.934 & & 0.190 & 1.153 & 4.802 & 0.287 \\
				\hline\hline
				\multirow{4}{*}{2D anchor BBox w/o distance prior} &\multirow{15}{*}{$<\boldsymbol{35}$}
				%& 2 & 0.896 & 0.976 & & 0.111 & 0.582 & 4.596 & 0.171  \\
				& 3 & 0.906 & 0.977 & & 0.103 & 0.547 & 4.475 & 0.167 \\
				&& 5 & 0.911 & 0.981 & & 0.096 & 0.474 & 4.225 & 0.157 \\
				&& 7 & 0.926 & 0.984 & & 0.085 & 0.410 & 3.727 & 0.145 \\
				\cline{1-1}\cline{3-10}
				\multirow{3}{*}{2D anchor BBox + average distance}
				%& & 2 & 0.905 & 0.980 & & 0.113 & 0.585 & 4.463 & 0.172  \\
				&& 3 & 0.914 & 0.982 & & 0.099 & 0.491 & 4.196 & 0.159 \\
				&& 5 & 0.923 & 0.984 & & 0.092 & 0.437 & 3.911 & 0.152 \\
				\cline{1-1}\cline{3-10}
				\multirow{4}{*}{Ours (normal anchor distance)}
				%&& 2 & 0.922 & 0.985 & & 0.103 & 0.478 & 4.142 & 0.158 \\
				&& 3 & 0.949 & 0.988 & & 0.094 & 0.344 & 3.401 & 0.144 \\
				&& 5 & 0.968 & 0.990 & & 0.084 & 0.230 & 2.527 & 0.131 \\
				&& 9 & 0.971 & 0.989 & & 0.076 & 0.155 & 1.770 & 0.124 \\
				\cline{1-1}\cline{3-10}
				\multirow{4}{*}{Ours (log-scale anchor distance)}
				%&& 2 & 0.906 & 0.983 & & 0.103 & 0.493 & 4.435 & 0.160 \\
				&& 3 & 0.931 & 0.987 & & 0.098 & 0.401 & 3.903 & 0.149 \\
				&& 5 & 0.957 & 0.989 & & 0.084 & 0.281 & 3.182 & 0.133 \\
				&& 9 & \textbf{0.972} & 0.990 & & \textbf{0.073} & 0.150 & 1.915 & 0.117 \\ 
				\cline{1-1}\cline{3-10}								  
				\multirow{4}{*}{Ours (squared anchor distance)}
				%&& 2 & 0.911 & 0.982 & & 0.109 & 0.521 & 4.173 & 0.164 \\
				&& 3 & 0.952 & 0.988 & & 0.092 & 0.313 & 2.936 & 0.142 \\
				&& 5 & 0.962 & 0.989 & & 0.084 & 0.220 & 2.080 & 0.134 \\
				&& 9 & 0.970 & 0.989 & & 0.079 & 0.165 & \textbf{1.719} & 0.127 \\
				\hline\hline
			\end{tabular}
		\end{center}
	\end{table*}
	\subsection{Depth Estimation and Anchor Distance}
	%object의 $z$ coordinate distance (depth) estimation에 대하여, 다양한 metric evaluation 결과를 Table.~\ref{distance_eval}에 나타내었다.
	We represent the metric evaluations of $z$ coordinate distance (depth) estimation of the object in Table~\ref{distance_eval}.
	We follow the definitions of the metrics as in \cite{zhu2019learning}.
	%Anchor distance의 유효성을 검증하기 위해, we also evaluate the methods of 2D anchor BBox with and without average distance.
	Compared to the methods based on RPN \cite{zhu2019learning,zhang2020regional}, our method achieves a better performance in RMSE at a substantially higher frame rate.
	Using depth estimation directly for object detection showed degraded performance compared to other methods as it hardly consider the overlapped or occluded states of objects.
	
	In order to validate the anchor distance, we also evaluate the methods of 2D anchor BBox with and without average distance.
	%섹션[method]에서도 언급하였듯이, anchor BBox와 average distance를 이용하는 경우 predictor가 추론해야 하는 distance의 variance가 커져서 추론하는 distance의 정확도가 떨어지는 것을 알 수 있다.
	%When using the anchor BBox and average distance, the variance of the distance to be inferred by one predictor increases.
	%그렇기 때문에 predictor의 수를 많이 할 수록 predictor 하나가 가지는 부담이 줄어들어 정확도가 증가한다.
	As the number of predictors increases, the burden of one predictor decreases and the network can achieve the robust estimations.
	%More the number of predictors are, more accurate inference is achieved.
	In other words, the more number of predictors there are, the more accurate inference is achieved.
	%이러한 경향은 anchor distance를 이용하였을 때 더욱 두드러지게 나타나며, 2D BBox에 집중하였을 때보다 더 좋은 성능을 보인다.
	This trend is more pronounced when anchor distance is applied, and in this case the network shows significantly improved performance than when the predictors simply focus on 2D BBox.
	%또한 predictor 수가 많아질 수록 더욱 정확한 추론이 가능하다.
	
	%제안하는 기법은 distance format이 squared일 때 가장 성능이 높은 것을 알 수 있으며, log-scale일때의 성능이 가장 낮다.
	The proposed method has the highest performance when the squared format is used, and shows the lowest performance when log-scale is used.
	We found that the squared distance format covers the largest range as shown in Table~\ref{anchor_distance_different_format}.
	With the large range of anchor distances, the network can handle the objects distributed in various ranges efficiently.
	\begin{table}[h]
		\caption{Anchor Distance of Different Distance Format}
		\label{anchor_distance_different_format}
		\begin{center}
			\begin{tabular}{c c c c c c c}
				& & & & & & (unit : $m$)\\
				\hline
				\hline
				\multicolumn{2}{c}{order of predictors} & 1  & 2 & 3 & 4 & 5 \\
				\hline
				\hline
				\multicolumn{2}{c}{anchor BBox(avr dist)} & 7.73 & 13.59 & 23.83 & 33.81 & 52.52 \\
				\hline
				\multirow{3}{*}{\specialcell{anchor\\ distance}} & normal & 11.20 & 23.18 & 35.30 & 49.50 & 66.21 \\
				& log-scale & 7.60 & 15.17 & 24.78 & 37.59 & 57.51 \\
				& squared & 17.49 & 32.86 & 45.27 & 57.41 & 71.52 \\
				\hline
				\hline
			\end{tabular}
		\end{center}
	\end{table}
	Meawhile, in the case of 2D BBox with average distance, prior distance is the most similar to that of log-scale format.
	Reversely, log-scale distance grouping shows the most similar average 2D BBox to that of 2D BBox grouping; we display the 2D BBoxes of each grouping in Table~\ref{BBox_compare}.
	\begin{table*}[t]
		\caption{Comparison of 2D BBox Dimensions for KITTI dataset}
		\label{BBox_compare}
		\begin{center}
			\begin{tabular}{c c cc c ccc c ccccc}
				&& & & & & & & & & & & & (unit : pixel)\\
				\hline
				\hline
				\# of predictors  && \multicolumn{2}{c}{2} && \multicolumn{3}{c}{3} && \multicolumn{5}{c}{5} \\
				\cline{1-1}\cline{3-4}\cline{6-8}\cline{10-14}
				orders && 1 & 2 && 1 & 2 & 3 && 1 & 2 & 3 & 4 & 5 \\
				\hline
				\hline
				anchor box(IoU)&& 151/268 & 51/86 && 164/296 & 83/144 & 39/64 && 175/321 & 108/183 & 57/104 & 37/61 & 24/35 \\
				anchor dist(log-scale)&& 139/226 & 37/67 && 156/246 & 59/103 & 29/53 && 181/273 & 104/178 & 57/99 & 36/67 & 23/41 \\
				anchor dist(normal)&& 133/219 & 30/55 && 141/229 & 42/75 & 24/42 && 156/246 & 63/111 & 38/69 & 27/49 & 19/32 \\
				anchor dist(squared)&& 129/215 & 24/43 && 134/220 & 33/60 & 21/36 && 140/227 & 42/75 & 29/54 & 22/39 & 17/29 \\
				\hline
				\hline
			\end{tabular}
		\end{center}
	\end{table*}
	Therefore, we can conclude that the distance in the log-scale format is mostly related to the 2D BBox, but using the log-scale distance or 2D BBox is not sufficient to achieve the best performance as the 2D BBox is not directly proportional to the distance.
	In other words, using log-scale distance or 2D BBox for grouping in order to obtain prior distance is not effective.
	For 3D localization tasks it is better concentrating not on 2D projection, but on distance itself for defining anchors of 3D location.
	
	\subsection{Estimation Error and Execution Time}
	The proposed method shows a significantly better performance on the RMSE error than that of others, but in the Abs Rel metric the method does not outperform the existing method using complex structures such as ResNet or Mask-RCNN.
	In other words, although our proposed method is slightly inferior to the previous methods for objects that are close to the origin, it is more robust when the objects are far from the origin.
	The structure of the proposed network is simpler than others, yet our approach using anchor distance and training scheme consistently generally achieves more accurate estimation regardless of the distance of objects.
	
	We also present the frame rate in terms of FPS for each method.
	In \cite{zhu2019learning}, they assume that the 2D BBox regression is given in advance, so we approximately record its FPS by assuming that YOLOv2 is used.
	In \cite{zhang2020regional}, the FPS of their method is under 7, as the baseline is MaskRCNN \cite{maskrcnn} that shows 7 FPS and the method deploys additional network structure for distance estimation.
	The proposed network, which is based on YOLOv2~\cite{redmon2017yolo9000}, however, only takes about 0.03 seconds and shows under 35 FPS for the entire estimation process.
	This execution time is irrelevant to the number of predictors, as adding one predictor is equivalent to adding $\left(4+1\right)$ convolutional filters that only have $\left(3\times3\right)$ parameters per filter.
	%Compared to the method that directly estimates all 3d coordinates, the methods using 2d BBox center shows stable estimation results for x coordinates(column-wise direction in 2d image plane).
	The method with 3D projection takes another 0.04 seconds after estimating 2D BBox, 3d BBox, and orientation which takes 0.03 seconds.
	\begin{figure}[t]
		\centering
		\includegraphics[scale=0.6]{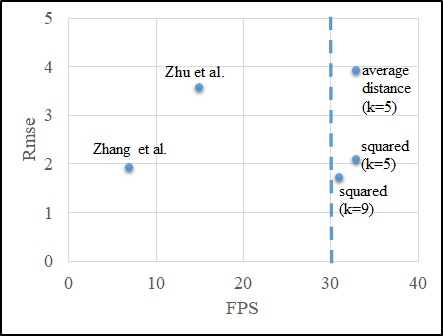}
		\caption{
			The RMSE and FPS of the methods are shown.
			Since our method is based on a real-time detector, it demonstrates high FPS.%
			Adopting the anchor distance as prior information compensates the simple network structure, achieving faster and better performance compared to others.%
		}%
		\label{RMSE_fps}%
	\end{figure}
	We display the relation between RMSE error and FPS of various methods in Fig.~\ref{RMSE_fps}.
	
\begin{figure*}[t]
		\centering
		\includegraphics[scale=0.215]{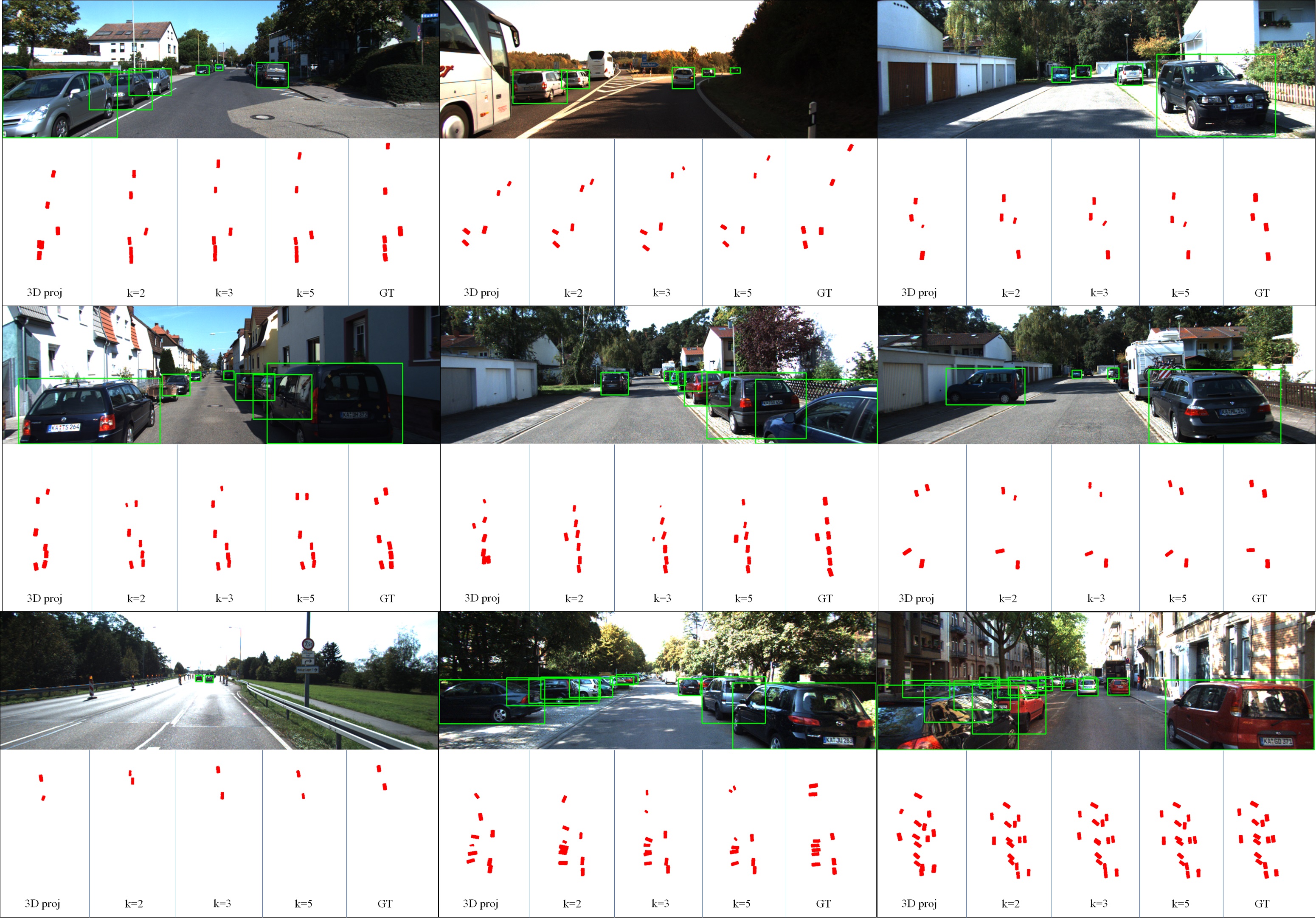}
		\caption{
			Examples of the visualizations in a bird-eye view.
			We compare the results of using 3D projection and the results of ours with $k=2,3,5$.
			Since the method using 3D projection highly depends on the 2D bounding box and orientation, it shows incorrect results for the objects located far away or occluded.
			The proposed method shows better performance with multiple predictors, as a number of anchor distances allows the network to estimate the results with low non-linear complexity.%
		}%
		\label{visualization}%
		\vspace{-7pt}
	\end{figure*}

	For qualitative analysis, we display the visualization examples in Fig.~\ref{visualization} represented in a bird-eye view.
	For \textit{3D proj}, we exploit the 2D BBox regression results from the network with $k=5$.
	The network for \textit{3D proj} shares the same structure of our proposed method.
	Also, it estimates orientations and dimensions of 2D and 3D BBox which are used for calculating distance.
	%제안하는 기법은 distance format이 squared인 경우를 이용하였다.
	Here, we use a squared format for the proposed method.
	%In 3D proj, 멀리 있는 물체일수록 2D BBox의 크기가 비슷해지기 때문에 거리 추정의 결과도 오차가 커진다.
	In \textit{3D proj}, the error of the distance estimation increases since the size of the 2D BBox becomes similar as the objects are located far away.
	%제안하는 기법은 the number of predictors $k$가 커질수록 추정한 결과가 정확해진다.
	In the proposed method, the higher the number of predictors $k$ is, the more accurate the estimated results are.

	\section{Conclusion}
	We propose a multi-object distance estimation method using anchor distance.
	%기존의 multi-object detection에 기반한 기법들은 2d BBox를 기준으로 하여 network의 predictor를 구분하고 학습하였다.
	Conventional methods based on multi-object detection train their predictors based on 2D Bounding Box (BBox).
	%multi-object의 distance추정을 위한 기법들도, 정교한 추정을 위해 복잡한 네트워크 구조를 기반으로 한다.
	Other techniques for multi-object distance estimation rely heavily on complex structure for sophisticated distance estimation.
	%제안하는 기법은 실시간으로 동작할 수 있는 네트워크로 물체의 거리를 정교하게 추정하기 위해, 물체의 distance를 기준으로 predictor를 골라 학습한다.
	The proposed method can achieve robust estimation and real-time performance as the method selects and trains the predictors of the network based on the distance of objects.
	%이를 위해 학습하고자 하는 물체를 다양한 format의 distance를 기준으로 clustering하여 anchor distance를 구하였다.
	To build a prior, we define the anchor distance by clustering the objects with their distance in various formats such as squared or log-scaled.
	%구한 anchor distance는 predictor에게 distance에 대한 강력한 prior을 제공한다.
	The anchor distance gives predictors a strong prior knowledge of distance.
	%predictor들은 anchor distance에 따라 특정 distance 범위에 있는 물체를 전담하여 학습하게 된다.
	Predictors are dedicated to learning objects in a specific distance range according to their anchor distances.
	%제안하는 방법은 distance를 기준으로 predictor를 학습시키게 되어 네트워크의 더욱 정확한 추정을 가능하게 한다.
	Because the proposed method trains the network based on distance, it is able to achieve more accurate estimations.
	%3d BBox를 2d BBox에 reprojection하여 distance를 계산하는 기존의 방법은 연산량이 많기 때문에 다수의 물체에 대하여 수행 속도가 제한된다.
	Traditional methods of directly calculating the distance by projecting 3D BBox to 2D BBox require a large amount of computation, so an increased number of objects tend to decline the speed of estimation during execution.
	%제안하는 기법은 real-time으로 동작하는 multi-object detector을 기반으로 하기 때문에 한 이미지당 물체의 개수에 관계없이 빠르게 distance를 estimate할 수 있다.
	%Since the proposed technique is based on real-time mult-object detector structure, it performs in real-time regardless of the number of objects in single shot.
	Using anchor distance as a prior the proposed approach develops a computationally concise network and performs single-shot, real-time, multi-object detection even for an arbitrarily large number of objects.
	%제안하는 기법은 물체의 localization만을 학습하였으나, 물체의 shape 및 dimensions of 3D BBox를 함께 학습하여 실시간 3D detection에도 적용할 수 있다.
	%The proposed method is designed for learning localizations of objects, but the concept of anchor distance can be applied to 3D object detection tasks.
	%이 경우 3D IoU 및 2D IoU loss를 추가할 수 있다.
	
	%{\small
	%	\bibliographystyle{ieee}
	%	\bibliography{bib_SLAM,bib_myPaper,bib_dataset}
	%}
	
	%\section*{Acknowledgement}
	%This work was supported by the National Research Foundation of Korea(NRF) grant funded by the Korea government(MSIP) (No. 2017R1A2B2002608), in part by Automation and Systems Research Institute (ASRI), and in part by the Brain Korea 21 Plus Project.
	
	\small{
		\section*{Acknowledgement}
		We would like to thank Jihoon Moon, who gives us intuitive advice.
		%This work was funded in part by the AI-Assisted Detection and Threat Recognition Program through the US ARMY ACC-APG-RTP under Contract No. W911NF1820218, ``Leveraging Advanced Algorithms, Autonomy, and Artificial Intelligence (A4I) to enhance National Security and Defense'' and the Air Force Office of Scientific Research under Award No. FA2386-17-1-4660.
		This work is in part supported by the Air Force Office of Scientific Research under award number FA2386-17-1-4660.
		%\clearpage
	}
	\bibliographystyle{IEEEtran}
	\bibliography{bib_my}
	
\end{document}